\documentclass[twoside,11pt]{article}

\usepackage[preprint]{jmlr2e}
\usepackage{tcolorbox}
\usepackage{color-edits}
\usepackage[scaled=0.8]{DejaVuSansMono}
\usepackage{enumitem}

\ShortHeadings{Fairlearn}{Weerts, Dud{\'i}k, Edgar, Jalali, Lutz, and Madaio}

\firstpageno{1}
\begin{document}
\title{Fairlearn: Assessing and Improving Fairness of AI Systems}

\author{\name Hilde Weerts\textsuperscript{1} \email h.j.p.weerts@tue.nl \\
        \name Miroslav Dud{\'i}k\textsuperscript{2} \email mdudik@microsoft.com \\
        \name Richard Edgar\textsuperscript{2} \email riedgar@microsoft.com \\
        \name Adrin Jalali
        \email adrin.jalali@gmail.com \\
        \name Roman Lutz\textsuperscript{2} \email romanlutz@microsoft.com \\
        \name Michael Madaio\textsuperscript{2}\thanks{Michael is currently employed by Google, but contributed to this work while at Microsoft.} \email madaiom@google.com \\
        \addr
            The authors are the current maintainers of Fairlearn, and additionally have the following affiliations:
        \\
        \addr
            \textsuperscript{1}Eindhoven University of Technology,
            \textsuperscript{2}Microsoft
       }

\editor{Editor Name}

\maketitle

\begin{abstract}
Fairlearn is an open source project to help practitioners assess and improve fairness of artificial intelligence (AI) systems. The associated Python library, also named \emph{fairlearn}, supports
evaluation of a model's output across affected populations and includes several algorithms for mitigating fairness issues. Grounded in the understanding that fairness is a sociotechnical challenge, the project integrates learning resources that aid practitioners in considering a system's broader societal context.
\end{abstract}


\defcitealias{ec2021proposal}{EC, 2021}
\defcitealias{ostp2022blueprint}{OSTP, 2022}

\section{Introduction}
As artificial intelligence (AI) impacts more of our everyday lives,
there is a growing need to ensure that algorithmic systems do not disproportionately harm minorities, historically disadvantaged populations, and other groups considered sensitive from an ethical or legal perspective
\citep{crawford2013hidden, o2016weapons, broussard2018artificial, noble2018algorithms, benjamin2019race}.
Fairness of AI systems is a topic of multiple academic venues,\footnote{%
  \url{https://facctconference.org/} (FAccT)\\
  \url{https://www.aies-conference.com/} (AIES)}
a priority for regulators~\citepalias{ec2021proposal,ostp2022blueprint},
and a focus of several open source projects~\citep{lee2021landscape}.

In this paper, we describe Fairlearn, an open source project that
seeks to help data science practitioners with assessing and improving fairness of AI systems.
The project consists of a Python library, called \emph{fairlearn}, accompanied with various learning resources. Both the library and the learning resources are licensed under MIT license
and available online.\footnote{%
  \url{https://github.com/fairlearn/fairlearn}\\
  \url{https://fairlearn.org}}
The library aims to provide an easy-to-use API that blends well with popular libraries of the Python ecosystem, including \emph{scikit-learn} \citep{scikit-learn}, \emph{pandas} \citep{pandas}, \emph{matplotlib} \citep{matplotlib}, \emph{TensorFlow}  \citep{tensorflow}, and \emph{PyTorch} \citep{pytorch}. Through our learning resources, we hope to provide practitioners with the knowledge and skills to effectively assess and mitigate unfairness.

Fairlearn is a community-driven project with independent governance,\footnote{%
The project started
in May 2018 as a Microsoft open source project and its initial scope was outlined in a Microsoft technical report~\citep{bird2020fairlearn}, but it transitioned to independent governance in~2021:\\\url{https://github.com/fairlearn/governance/blob/main/ORG-GOVERNANCE.md}}
following a code of conduct adapted from the Contributor Covenant.\footnote{\url{https://www.contributor-covenant.org}}
The project is under active development and welcomes community contributions to the source code and the learning resources.

\paragraph{Fairlearn Perspective on AI Fairness.}

In Fairlearn, we consider AI fairness through the lens of fairness-related harms~\citep{crawford2017trouble}, by which we mean negative impacts for groups of people, such as those defined in terms of race, gender, age or disability status.

Development of Fairlearn is firmly grounded in the understanding that fairness of AI systems is a sociotechnical challenge \cite[cf.][]{green2021contestation}. Because there are many complex sources of unfairness---some societal and some technical---it is not possible to fully ``de-bias'' a system or to guarantee fairness \cite[e.g.,][]{blodgett2020language}. Instead, our goal is to help practitioners assess fairness-related harms, review the impacts of different mitigation strategies, and make trade-offs appropriate to their scenario. This may sometimes mean advocating for not deploying the system at all \citep{baumer2011implication}. AI fairness is related to, but distinct from anti-discrimination laws~\citep{xiang2019legal}, so our documentation avoids (mis)use of legal terminology \citep{watkins2022four} and encourages users to understand what fairness means for their sociotechnical context before applying or adapting Fairlearn.\looseness=-1

Fairlearn largely focuses on two types of fairness-related harms: \emph{allocation harms} and \emph{quality-of-service harms}.
Allocation harms occur when AI systems are used to allocate opportunities or resources in ways that can have significant negative impacts on people's lives, for example,
when an AI system for recommending patients into high-risk care management programs is less likely to select Black patients than white patients of similar health~\citep{obermeyer2019dissecting}. Quality-of-service harms occur when a system does not work as well for members of one group as it does for members of another group, for example, when a computer vision system has higher error rates for images of women with darker skin than for images of men with lighter skin~\citep{buolamwini18gender}.

\section{Fairness Assessment}

One of the key goals of the \emph{fairlearn} library is to support fairness assessment.
The goal of fairness assessment is to answer the question: \emph{Which groups of people may be disproportionately negatively impacted by an AI system and in what ways?}
In the context of allocation and quality-of-service harms, this means to evaluate how well the system performs for different population groups by calculating some performance metric, like an error rate, on different slices of data. This is called \emph{disaggregated evaluation}~\citep{barocas2021designing}.

\paragraph{\texttt{MetricFrame} class.}
The primary tool for disaggregated evaluation in the \emph{fairlearn} library is the \path{MetricFrame} class in the \path{fairlearn.metrics} module. Its API combines \emph{scikit-learn} and \emph{pandas} conventions. In its simplest form, \path{MetricFrame} is initialized by providing one or more metric functions together with input arrays \path{y_true}, \path{y_pred}, and \path{sensitive_features}. The first two arrays serve as inputs to metric functions, whereas the \path{sensitive_features} array is used to split the data into slices for disaggregated evaluation.
Once a \path{MetricFrame} is constructed, the disaggregated metrics can be accessed as a \emph{pandas} \path{Series} (for a single metric) or a \emph{pandas} \path{DataFrame} (if multiple metrics are provided).
\path{MetricFrame} also enables a comparison of metric values across groups, for example, in terms of differences or ratios. Plotting of results is supported via existing integration of \emph{pandas} with \emph{matplotlib}.

\paragraph{Fairness Metrics.}

The module \path{fairlearn.metrics} also provides metric functions that return scalars much like typical \emph{scikit-learn} metrics.
For example, functions \path{demographic_parity_difference} and \path{equalized_odds_difference} quantify how much
the predictions of a given classifier depart from the fairness criteria known as \emph{demographic parity} and \emph{equalized odds} \citep[see, e.g.,][]{hardt2016equality}. These two metrics are derived from a \path{MetricFrame} with a specific choice of input arguments.
New fairness metrics can be obtained by using the \path{make_derived_metric} function, which wraps some of the \path{MetricFrame} functionality.

\paragraph{Comparison of Multiple Models.}

In addition to assessing fairness of a single model, \path{fairlearn.metrics} also enables a comparison of multiple models. For example, the function \path{plot_model_comparison} can be used to create a scatter plot, where each model is represented as a point with one coordinate equal to a metric quantifying overall performance and the other to a metric quantifying fairness, like the metrics from the \path{fairlearn.metrics} module.

\section{Algorithmic Mitigation of Fairness-related Harms}

The \emph{fairlearn} library includes several methods for mitigating fairness-related
harms. Many of the included methods are \emph{meta-algorithms} in the sense that they act as wrappers around any standard (i.e., fairness-unaware) machine learning algorithms. This makes them quite versatile in practice. All of the implementations follow
the API conventions of \emph{scikit-learn}.

Following \citet{barocas-hardt-narayanan}, \emph{fairlearn} mitigation algorithms can be divided into three groups according to when they are applied relative to model training:

\paragraph{Pre-processing.}
Algorithms in this group mitigate unfairness by transforming input data before it is passed to a standard training algorithm.
For example, \path{CorrelationRemover} in the module \path{fairlearn.preprocessing} applies
a linear transformation to input features in order to remove any correlation with sensitive features. It follows the API
of a \emph{scikit-learn} transformer and therefore can be incorporated in a \emph{scikit-learn} pipeline.

\paragraph{In-training.\protect\footnote{Also called \emph{in-processing} by some authors~\citep{kamiran2013techniques}.}}
Algorithms in this group directly train a model to satisfy fairness constraints.
For example, the meta-algorithm \path{ExponentiatedGradient} in the module \path{fairlearn.reductions} implements
the reduction approach of \citet{agarwal2018reductions,agarwal2019fair}. This meta-algorithm supports a wide range of fairness constraints and wraps any standard classification or regression algorithm, such as \path{LogisticRegression} from \path{sklearn.linear_model} or \path{XGBRegressor} from \path{xgboost}. An input to a reduction algorithm is an object that supports training on any provided (weighted) data set as well as a data set that includes sensitive features. The goal is to optimize a performance metric (such as classification accuracy) subject to fairness constraints (such as an upper bound on a difference between false negative rates).\looseness=-1

As another example, \path{AdversarialClassifier} and \path{AdversarialRegressor} in the module
\path{fairlearn.adversarial} implement the adversarial mitigation approach of \citet{zhang2018mitigating}. These algorithms simultaneously train two neural network models, a predictor model and an adversarial model. The predictor model seeks to minimize the prediction loss function while also ensuring that the adversary model cannot infer sensitive features from the predictor outputs.
The predictor and adversary neural nets can be defined either as \emph{PyTorch} modules or \emph{TensorFlow} models.

\paragraph{Post-processing.}
Algorithms in this group transform the output of a trained model.
For example,
\path{ThresholdOptimizer} in the module \path{fairlearn.postprocessing} implements the approach of \citet{hardt2016equality}, which takes in an existing (possibly pre-fit) machine learning model, uses its predictions as a scoring function, and identifies a separate threshold for each group defined by a sensitive feature in order to optimize some specified objective (such as balanced accuracy) subject to specified fairness constraints (such as false negative rate parity). The resulting classifier is a thresholded version of the provided machine learning model.\looseness=-1

\section{Learning Resources}

Tackling fairness-related harms requires more than technical tools alone \citep{holstein2019improving}. In a community-based effort, we have developed a comprehensive set of \textit{learning objectives}
that highlight what practitioners should know or be able to do when assessing and improving fairness of AI systems. These objectives are the basis for our learning resources.\looseness=-1

To avoid divorcing technical and social aspects of AI fairness, our learning resources are integrated
in the API reference and user guide of the \emph{fairlearn} library. Besides coding examples and explanations, our user guide covers important concepts central to understanding machine learning models as part of a sociotechnical system, such as construct validity~\citep{jacobs2021measurement} and the risks of abstracting away social context~\citep{selbst2019fairness}.\looseness=-1

Examples are crucial when learning to view fairness from a sociotechnical perspective. We provide tutorials~\citep[e.g.,][]{scipytutorial2021} and example notebooks downloadable in the Jupyter format~\citep{kluyver2016jupyter}. We try to ensure that each notebook describes a real-world or realistic deployment context, focuses on real harms to real people, and avoids \textit{abstraction traps}~\citep{selbst2019fairness}.

The data sets provided in the module \path{fairlearn.datasets} also serve an educational role, as we use them to highlight sociotechnical aspects of fairness, with sections of the user guide highlighting fairness-related issues with several popular benchmark data sets.

\section{Conclusions}

Fairlearn is built and maintained by contributors with a variety of backgrounds and expertise. We believe that meaningful progress toward fairer AI systems
requires input from a breadth of perspectives. We therefore encourage researchers,
practitioners, and other stakeholders to contribute to Fairlearn as we experiment, learn, and evolve the project together.\looseness=-1

\newpage
\acks{We would like to thank
Sarah Bird, Brandon Horn, Vanessa Milan, Mehrnoosh Sameki, Hanna Wallach, and Kathleen Walker
for their critical contributions to the initial Fairlearn project~\citep{bird2020fairlearn}.
We would also like to thank all the members of the Fairlearn community who have contributed to the project in various ways, including documentation, code, bug reports, feature requests, and participation in our community calls. In particular, we would like to thank Michael Amoako, Alexandra Chouldechova, Parul Gupta, Laura Gutierrez Funderburk, Abdul Hannan Kanji, Kenneth Holstein, Lisa Iba\~nez, Sean McCarren, Manojit Nandi, Ayodele Odubela, Rens Oostenbach, Alex Quach, Kevin Robinson, Allie Saizan, Bram Schut, and Vincent Warmerdam for their valuable contributions.\looseness=-1}

\bibliography{sample}

\begin{thebibliography}{34}
\providecommand{\natexlab}[1]{#1}
\providecommand{\url}[1]{\texttt{#1}}
\expandafter\ifx\csname urlstyle\endcsname\relax
  \providecommand{\doi}[1]{doi: #1}\else
  \providecommand{\doi}{doi: \begingroup \urlstyle{rm}\Url}\fi

\bibitem[Abadi et~al.(2015)Abadi, Agarwal, Barham, Brevdo, Chen, Citro,
  Corrado, Davis, Dean, Devin, Ghemawat, Goodfellow, Harp, Irving, Isard, Jia,
  Jozefowicz, Kaiser, Kudlur, Levenberg, Man\'{e}, Monga, Moore, Murray, Olah,
  Schuster, Shlens, Steiner, Sutskever, Talwar, Tucker, Vanhoucke, Vasudevan,
  Vi\'{e}gas, Vinyals, Warden, Wattenberg, Wicke, Yu, and Zheng]{tensorflow}
Mart\'{i}n Abadi, Ashish Agarwal, Paul Barham, Eugene Brevdo, Zhifeng Chen,
  Craig Citro, Greg~S. Corrado, Andy Davis, Jeffrey Dean, Matthieu Devin,
  Sanjay Ghemawat, Ian Goodfellow, Andrew Harp, Geoffrey Irving, Michael Isard,
  Yangqing Jia, Rafal Jozefowicz, Lukasz Kaiser, Manjunath Kudlur, Josh
  Levenberg, Dandelion Man\'{e}, Rajat Monga, Sherry Moore, Derek Murray, Chris
  Olah, Mike Schuster, Jonathon Shlens, Benoit Steiner, Ilya Sutskever, Kunal
  Talwar, Paul Tucker, Vincent Vanhoucke, Vijay Vasudevan, Fernanda Vi\'{e}gas,
  Oriol Vinyals, Pete Warden, Martin Wattenberg, Martin Wicke, Yuan Yu, and
  Xiaoqiang Zheng.
\newblock {TensorFlow}: Large-scale machine learning on heterogeneous systems,
  2015.
\newblock Software available from \url{https://www.tensorflow.org}.

\bibitem[Agarwal et~al.(2018)Agarwal, Beygelzimer, Dud{\'\i}k, Langford, and
  Wallach]{agarwal2018reductions}
Alekh Agarwal, Alina Beygelzimer, Miroslav Dud{\'\i}k, John Langford, and Hanna
  Wallach.
\newblock {A Reductions Approach to Fair Classification}.
\newblock In \emph{International Conference on Machine Learning}, pages 60--69,
  2018.

\bibitem[Agarwal et~al.(2019)Agarwal, Dud\'ik, and Wu]{agarwal2019fair}
Alekh Agarwal, Miroslav Dud\'ik, and Zhiwei~Steven Wu.
\newblock {Fair Regression: Quantitative Definitions and Reduction-Based
  Algorithms}.
\newblock In \emph{International Conference on Machine Learning}, pages
  120--129, 2019.

\bibitem[Barocas et~al.(2019)Barocas, Hardt, and
  Narayanan]{barocas-hardt-narayanan}
Solon Barocas, Moritz Hardt, and Arvind Narayanan.
\newblock \emph{Fairness and Machine Learning: Limitations and Opportunities}.
\newblock 2019.
\newblock \url{https://www.fairmlbook.org}.

\bibitem[Barocas et~al.(2021)Barocas, Guo, Kamar, Krones, Morris, Vaughan,
  Wadsworth, and Wallach]{barocas2021designing}
Solon Barocas, Anhong Guo, Ece Kamar, Jacquelyn Krones, Meredith~Ringel Morris,
  Jennifer~Wortman Vaughan, W.~Duncan Wadsworth, and Hanna Wallach.
\newblock {Designing Disaggregated Evaluations of AI Systems: Choices,
  Considerations, and Tradeoffs}.
\newblock In \emph{Proceedings of the 2021 AAAI/ACM Conference on AI, Ethics,
  and Society}, AIES '21, pages 368--378, 2021.

\bibitem[Baumer and Silberman(2011)]{baumer2011implication}
Eric~PS Baumer and M~Six Silberman.
\newblock When the implication is not to design (technology).
\newblock In \emph{Proceedings of the SIGCHI Conference on Human Factors in
  Computing Systems}, pages 2271--2274, 2011.

\bibitem[Benjamin(2019)]{benjamin2019race}
Ruha Benjamin.
\newblock \emph{Race After Technology: Abolitionist Tools for the New Jim
  Code}.
\newblock Polity, 2019.

\bibitem[Bird et~al.(2020)Bird, Dud{\'i}k, Edgar, Horn, Lutz, Milan, Sameki,
  Wallach, and Walker]{bird2020fairlearn}
Sarah Bird, Miroslav Dud{\'i}k, Richard Edgar, Brandon Horn, Roman Lutz,
  Vanessa Milan, Mehrnoosh Sameki, Hanna Wallach, and Kathleen Walker.
\newblock {Fairlearn: A Toolkit for Assessing and Improving Fairness in {AI}}.
\newblock Technical Report MSR-TR-2020-32, Microsoft, May 2020.
\newblock \url{https://aka.ms/fairlearn-whitepaper}.

\bibitem[Blodgett et~al.(2020)Blodgett, Barocas, Daum{\'e}~III, and
  Wallach]{blodgett2020language}
Su~Lin Blodgett, Solon Barocas, Hal Daum{\'e}~III, and Hanna Wallach.
\newblock Language (technology) is power: A critical survey of" bias" in nlp.
\newblock \emph{arXiv preprint arXiv:2005.14050}, 2020.

\bibitem[Broussard(2018)]{broussard2018artificial}
Meredith Broussard.
\newblock \emph{{Artificial Unintelligence: How Computers Misunderstand the
  World}}.
\newblock MIT Press, 2018.

\bibitem[Buolamwini and Gebru(2018)]{buolamwini18gender}
Joy Buolamwini and Timnit Gebru.
\newblock {Gender Shades: {I}ntersectional Accuracy Disparities in Commercial
  Gender Classification}.
\newblock In \emph{Proceedings of the 1st Conference on Fairness,
  Accountability and Transparency}, pages 77--91, 2018.

\bibitem[Crawford(2013)]{crawford2013hidden}
Kate Crawford.
\newblock {The Hidden Biases in Big Data}.
\newblock \emph{Harvard business review}, 1\penalty0 (4), 2013.

\bibitem[Crawford(2017)]{crawford2017trouble}
Kate Crawford.
\newblock {The Trouble with Bias}.
\newblock NeurIPS keynote, 2017.
\newblock \url{https://www.youtube.com/watch?v=fMym_BKWQzk}.

\bibitem[{(EC)}(2021)]{ec2021proposal}
{European Commission} {(EC)}.
\newblock Proposal for a {R}egulation of the {E}uropean {P}arliament and of the
  {C}ouncil laying down harmonised rules on artificial intelligence
  {(Artificial Intelligence Act)} and amending certain {U}nion legislative
  acts, 2021.
\newblock COM (2021) 206 final.

\bibitem[Gandhi et~al.(2021)Gandhi, Nandi, Dud{\'\i}k, Wallach, Madaio, Weerts,
  Jalali, and Iba{\~n}ez]{scipytutorial2021}
Triveni Gandhi, Manojit Nandi, Miroslav Dud{\'\i}k, Hanna Wallach, Michael
  Madaio, Hilde Weerts, Adrin Jalali, and Lisa Iba{\~n}ez.
\newblock {Fairness in {AI} Systems: From Social Context to Practice using
  Fairlearn}.
\newblock Tutorial presented at the 20th annual Scientific Computing with
  Python Conference (Scipy 2021), Virtual Event, 2021.

\bibitem[Green(2021)]{green2021contestation}
Ben Green.
\newblock The contestation of tech ethics: A sociotechnical approach to
  technology ethics in practice.
\newblock \emph{Journal of Social Computing}, 2\penalty0 (3):\penalty0
  209--225, 2021.

\bibitem[Hardt et~al.(2016)Hardt, Price, Price, and Srebro]{hardt2016equality}
Moritz Hardt, Eric Price, Eric Price, and Nati Srebro.
\newblock {Equality of Opportunity in Supervised Learning}.
\newblock In \emph{Advances in Neural Information Processing Systems}, 2016.
\newblock \url{https://arxiv.org/abs/1610.02413}.

\bibitem[Holstein et~al.(2019)Holstein, Wortman~Vaughan, Daum\'{e}, Dud\'ik,
  and Wallach]{holstein2019improving}
Kenneth Holstein, Jennifer Wortman~Vaughan, Hal Daum\'{e}, Miroslav Dud\'ik,
  and Hanna Wallach.
\newblock Improving fairness in machine learning systems: What do industry
  practitioners need?
\newblock In \emph{Proceedings of the 2019 CHI Conference on Human Factors in
  Computing Systems}, CHI '19, pages 1--16, 2019.

\bibitem[Hunter(2007)]{matplotlib}
J.~D. Hunter.
\newblock {Matplotlib: A 2D graphics environment}.
\newblock \emph{Computing in Science \& Engineering}, 9\penalty0 (3):\penalty0
  90--95, 2007.
\newblock \doi{10.1109/MCSE.2007.55}.

\bibitem[Jacobs and Wallach(2021)]{jacobs2021measurement}
Abigail~Z. Jacobs and Hanna Wallach.
\newblock {Measurement and Fairness}.
\newblock In \emph{Proceedings of the 2021 ACM Conference on Fairness,
  Accountability, and Transparency}, FAccT '21, pages 375--–385, 2021.

\bibitem[Kamiran et~al.(2013)Kamiran, Calders, and
  Pechenizkiy]{kamiran2013techniques}
Faisal Kamiran, Toon Calders, and Mykola Pechenizkiy.
\newblock Techniques for discrimination-free predictive models.
\newblock In Bart Custers, Toon Calders, Bart~W. Schermer, and Tal~Z. Zarsky,
  editors, \emph{Discrimination and Privacy in the Information Society - Data
  Mining and Profiling in Large Databases}, volume~3 of \emph{Studies in
  Applied Philosophy, Epistemology and Rational Ethics}, pages 223--239.
  Springer, 2013.
\newblock \doi{10.1007/978-3-642-30487-3\_12}.
\newblock URL \url{https://doi.org/10.1007/978-3-642-30487-3\_12}.

\bibitem[Kluyver et~al.(2016)Kluyver, Ragan-Kelley, P{\'e}rez, Granger,
  Bussonnier, Frederic, Kelley, Hamrick, Grout, Corlay, Ivanov, Avila, Abdalla,
  and Willing]{kluyver2016jupyter}
Thomas Kluyver, Benjamin Ragan-Kelley, Fernando P{\'e}rez, Brian Granger,
  Matthias Bussonnier, Jonathan Frederic, Kyle Kelley, Jessica Hamrick, Jason
  Grout, Sylvain Corlay, Paul Ivanov, Dami{\'a}n Avila, Safia Abdalla, and
  Carol Willing.
\newblock {Jupyter Notebooks -- A Publishing Format for Reproducible
  Computational Workflows}.
\newblock In F.~Loizides and B.~Schmidt, editors, \emph{Positioning and Power
  in Academic Publishing: Players, Agents and Agendas}, pages 87--90. IOS
  Press, 2016.

\bibitem[Lee and Singh(2021)]{lee2021landscape}
Michelle Seng~Ah Lee and Jat Singh.
\newblock {The Landscape and Gaps in Open Source Fairness Toolkits}.
\newblock In \emph{Proceedings of the 2021 CHI Conference on Human Factors in
  Computing Systems}, CHI '21, 2021.

\bibitem[McKinney(2010)]{pandas}
Wes McKinney.
\newblock { Data Structures for Statistical Computing in {P}ython }.
\newblock In \emph{Proceedings of the 9th Python in Science Conference}, pages
  56--61, 2010.

\bibitem[Noble(2018)]{noble2018algorithms}
Safiya~Umoja Noble.
\newblock \emph{Algorithms of Oppression: How Search Engines Reinforce Racism}.
\newblock NYU Press, 2018.

\bibitem[Obermeyer et~al.(2019)Obermeyer, Powers, Vogeli, and
  Mullainathan]{obermeyer2019dissecting}
Ziad Obermeyer, Brian Powers, Christine Vogeli, and Sendhil Mullainathan.
\newblock {Dissecting Racial Bias in an Algorithm Used to Manage the Health of
  Populations}.
\newblock \emph{Science}, 366\penalty0 (6464):\penalty0 447--453, 2019.

\bibitem[O'Neil(2016)]{o2016weapons}
Cathy O'Neil.
\newblock \emph{Weapons of Math Destruction: How Big Data Increases Inequality
  and Threatens Democracy}.
\newblock Crown, 2016.

\bibitem[{(OSTP)}(2022)]{ostp2022blueprint}
{White House Office of Science and Technology Policy} {(OSTP)}.
\newblock {A Blueprint for an {AI} Bill of Rights}, 2022.
\newblock \url{https://www.whitehouse.gov/ostp/ai-bill-of-rights/}.

\bibitem[Paszke et~al.(2019)Paszke, Gross, Massa, Lerer, Bradbury, Chanan,
  Killeen, Lin, Gimelshein, Antiga, Desmaison, Kopf, Yang, DeVito, Raison,
  Tejani, Chilamkurthy, Steiner, Fang, Bai, and Chintala]{pytorch}
Adam Paszke, Sam Gross, Francisco Massa, Adam Lerer, James Bradbury, Gregory
  Chanan, Trevor Killeen, Zeming Lin, Natalia Gimelshein, Luca Antiga, Alban
  Desmaison, Andreas Kopf, Edward Yang, Zachary DeVito, Martin Raison, Alykhan
  Tejani, Sasank Chilamkurthy, Benoit Steiner, Lu~Fang, Junjie Bai, and Soumith
  Chintala.
\newblock {PyTorch: An Imperative Style, High-Performance Deep Learning
  Library}.
\newblock In \emph{Advances in Neural Information Processing Systems}, 2019.

\bibitem[Pedregosa et~al.(2011)Pedregosa, Varoquaux, Gramfort, Michel, Thirion,
  Grisel, Blondel, Prettenhofer, Weiss, Dubourg, Vanderplas, Passos,
  Cournapeau, Brucher, Perrot, and Duchesnay]{scikit-learn}
F.~Pedregosa, G.~Varoquaux, A.~Gramfort, V.~Michel, B.~Thirion, O.~Grisel,
  M.~Blondel, P.~Prettenhofer, R.~Weiss, V.~Dubourg, J.~Vanderplas, A.~Passos,
  D.~Cournapeau, M.~Brucher, M.~Perrot, and E.~Duchesnay.
\newblock Scikit-learn: Machine learning in {P}ython.
\newblock \emph{Journal of Machine Learning Research}, 12:\penalty0 2825--2830,
  2011.

\bibitem[Selbst et~al.(2019)Selbst, {boyd}, Friedler, Venkatasubramanian, and
  Vertesi]{selbst2019fairness}
Andrew~D Selbst, {danah} {boyd}, Sorelle~A Friedler, Suresh Venkatasubramanian,
  and Janet Vertesi.
\newblock {Fairness and Abstraction in Sociotechnical Systems}.
\newblock In \emph{Proceedings of the conference on fairness, accountability,
  and transparency}, pages 59--68, 2019.

\bibitem[Watkins et~al.(2022)Watkins, McKenna, and Chen]{watkins2022four}
Elizabeth~Anne Watkins, Michael McKenna, and Jiahao Chen.
\newblock The four-fifths rule is not disparate impact: a woeful tale of
  epistemic trespassing in algorithmic fairness.
\newblock \emph{arXiv preprint arXiv:2202.09519}, 2022.

\bibitem[Xiang and Raji(2019)]{xiang2019legal}
Alice Xiang and Inioluwa~Deborah Raji.
\newblock {On the Legal Compatibility of Fairness Definitions}.
\newblock Workshop on Human-Centric Machine Learning at NeurIPS, 2019.
\newblock \url{https://arxiv.org/abs/1912.00761}.

\bibitem[Zhang et~al.(2018)Zhang, Lemoine, and Mitchell]{zhang2018mitigating}
Brian~Hu Zhang, Blake Lemoine, and Margaret Mitchell.
\newblock {Mitigating Unwanted Biases with Adversarial Learning}.
\newblock In \emph{Proceedings of the 2018 AAAI/ACM Conference on AI, Ethics,
  and Society}, pages 335--340, 2018.

\end{thebibliography}

\end{document}